\documentclass[letterpaper, 10 pt, conference]{ieeeconf}  

\usepackage{etoolbox}
\makeatletter
\patchcmd{\@makecaption}
  {\scshape}
  {}
  {}
  {}
\makeatletter
\patchcmd{\@makecaption}
  {\\}
  {.\ }
  {}
  {}
\makeatother

\IEEEoverridecommandlockouts                              

\usepackage{graphicx} 
\usepackage{subfigure}

\usepackage{cite}
\usepackage{color}
\usepackage{float}
\usepackage{amsmath}
\usepackage{amssymb}
\usepackage{url}

\usepackage{capt-of}
\usepackage{algorithm}
\usepackage{algorithmic}

\begin{document}

\title{\LARGE 
One-Shot Hierarchical Imitation Learning of Compound Visuomotor Tasks
}

\author{Tianhe Yu$^{1}$, Pieter Abbeel$^{1}$, Sergey Levine$^{1}$, Chelsea Finn$^{1}$
\thanks{$^{1}$ Berkeley AI Research, UC Berkeley Computer Science. Correspondence to
        {\tt\small tianheyu@cs.stanford.edu}}%
        }

\maketitle

\begin{abstract}
We consider the problem of learning multi-stage vision-based tasks on a real robot from a single video of a human performing the task, while leveraging demonstration data of subtasks with other objects. This problem presents a number of major challenges. Video demonstrations without teleoperation are easy for humans to provide, but do not provide any direct supervision. Learning  policies from raw pixels enables full generality but calls for large function approximators with many parameters to be learned. Finally, compound tasks can require impractical amounts of demonstration data, when treated as a monolithic skill. To address these challenges, we propose a method that learns both how to learn primitive behaviors from video demonstrations and how to dynamically compose these behaviors to perform multi-stage tasks by ``watching'' a human demonstrator. Our results on a simulated Sawyer robot and real PR2 robot illustrate our method for learning a variety of order fulfillment and kitchen serving tasks with novel objects and raw pixel inputs. Video results are linked at \url{https://sites.google.com/view/one-shot-hil}.
\end{abstract}

\section{Introduction}
\label{sec:intro}


Humans have a remarkable ability to imitate complex behaviors from just \emph{watching} another person. In contrast, robots require a human to physically guide or teleoperate the robot's body~\cite{pastor_kinesthetic,akgun2012trajectories,hapticteleop_lfd,florida_vision,vr_imitation} in order to learn from demonstrations.
Furthermore, people can effectively learn behaviors from just a single demonstration, while robots often require substantially more data~\cite{florida_lstm,vr_imitation}. While prior work has made progress on imitating manipulation primitives using raw video demonstrations from humans~\cite{abhishek,pierre_time,daml_rss,third_person},
handling more complex, compound tasks presents an additional challenge. When skill sequences become temporally extended, it becomes impractical to treat a sequence of skills as a single, monolithic task, as the full sequence has a much longer time horizon than individual primitives and learning requires significantly more data.
In this paper, we consider the following question: can we leverage the compositional structure underlying compound tasks to effectively learn temporally-extended tasks from a single video of a human demonstrator?

\begin{figure}[t]
\centering
\includegraphics[width=1.0\columnwidth]{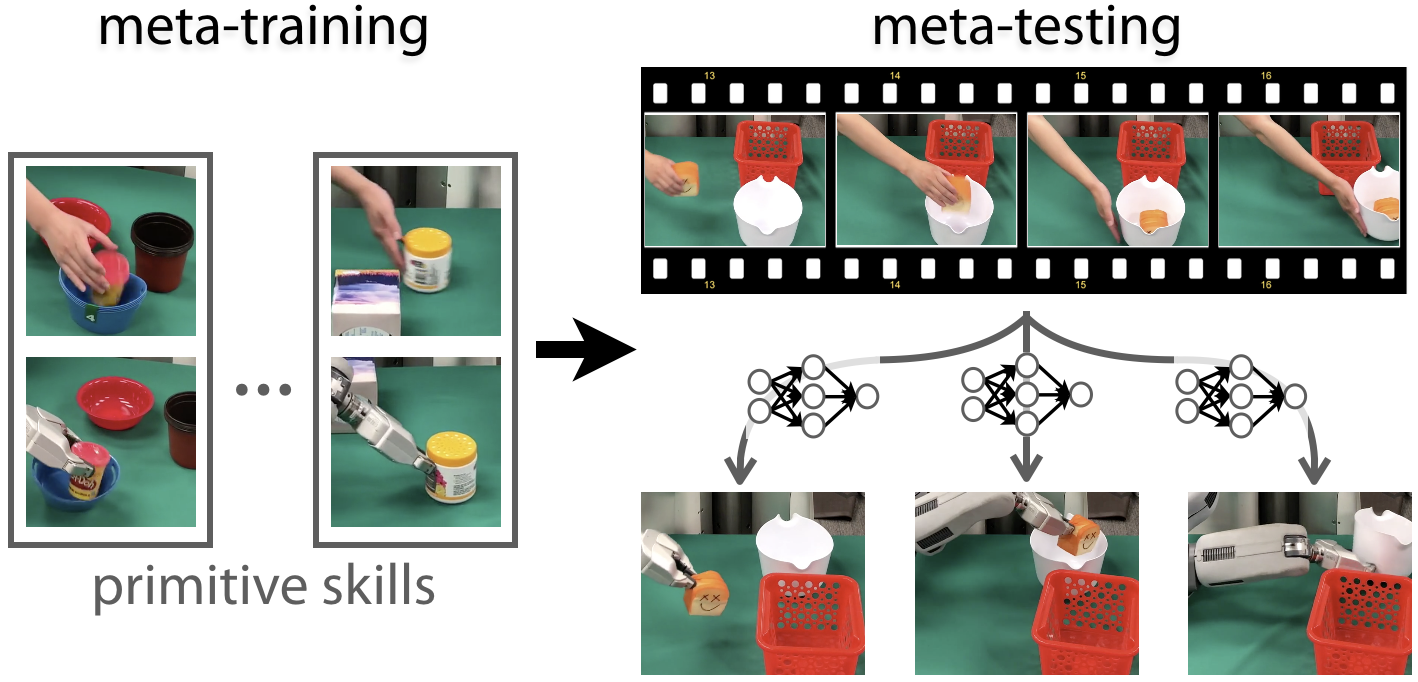}
\vspace{-0.6cm}
\caption{
\label{fig:teaser}
A robot learns and composes convolutional neural network policies for performing a multi-stage manipulation task by ``watching'' a human perform the compound task (right), by leveraging data from primitive skills with other objects (left).}
\vspace{-0.24in}
\end{figure}

A number of prior works aim to learn temporally extended skills using demonstration data that is labeled based on the particular primitive executed or using pre-programmed primitives~\cite{manschitz2015learning,maryland,xu2017neural,shiarlis2018taco}. 
However, when the human demonstrations are provided as raw videos, and robot must also handle raw visual observations, it is difficult to employ conventionally methods such as low-dimensional policy representations~\cite{contextdmp, adjustdmp} or changepoint detection~\cite{konidaris2012robot}. In this paper, we consider a problem setting of learning to perform multi-stage tasks through imitation where the robot must map raw image observations to actions, and the demonstration is provided via an unsegmented raw video of a person performing the entire task. To approach this problem, the key idea in this work is to leverage meta-learning, where the robot uses \emph{previous} data of primitive skills to learn how to imitate humans performing multi-stage skills. In particular, the meta-training set of previous \emph{primitive} skills consists of both videos of humans and teleoperated episodes, while the new \emph{multi-stage} tasks seen at meta-test time are only specified with a single video of a human, without teleoperation (see Figure~\ref{fig:teaser}).
Hence, our goal is to both learn primitive behaviors and to compose them automatically from a single video of a human performing the new compound task.

Our approach adapts its policy online over the course of the multi-stage task as it ``watches'' the video of the human. To accomplish this, we build models that can recognize the progress towards completion of the current primitive (or the primitive's `phase'), and integrate these models with a one-shot imitator to dynamically learn and compose primitives into compound tasks, from a video of a human. The phase of a primitive can be learned directly from the demonstration data of primitives, using the frame indices of each demonstration, without requiring any manual labeling. To learn a policy for an individual primitive from a video of a human performing that primitive, a building block of our approach, we use domain-adaptive meta-imitation learning~\cite{daml_rss}. All in all, as illustrated in Figure~\ref{fig:diagram}, our method decomposes the test-time human video into primitives using a primitive phase predictor, computes a sequence of policies for each primitive, and sequentially executes each policy until each has deemed it is complete, again utilizing a phase predictor.

The primary contribution of this work is an approach for  dynamically learning and composing sequences of policies based on a single human demonstration without annotations. 
Our method combines one-shot imitation of subtasks with a learned mechanism for decomposing compound task demonstrations and composing primitive skills.
In our experimental results, we show that this approach can be used to dynamically learn
and sequence skills from a single video demonstration provided by the user at test time.  We evaluate our method on order fulfillment tasks with a simulated Sawyer arm and a kitchen serving tasks with a real PR2 robot.
Our method learns to perform temporally-extended tasks from raw pixel inputs and outperforms alternative approaches.

\section{Related Work}
\label{related}

Prior methods for compound task learning from demonstrations often use demonstrations that are labeled or segmented into individual activities~\cite{maryland}, commands~\cite{rnn_translation,xu2017neural,shiarlis2018taco}, or textual descriptions~\cite{aksoy2017unsupervised}, or assume a library of pre-specified primitives~\cite{meier2011movement,pastor2012towards}. Unlike many of these approaches, we use a dataset of demonstrations that are not labeled by activity or command, reducing the burden on labeling. Further, at test time, we translate directly from a human video to a sequence of policies, bypassing the intermediate grounding in language or activity labels. This ``end-to-end'' approach has the benefit of making it straight-forward to convey motions or subtasks that are difficult to describe, such as the transitions from one subtask to another; it also does not require the human to have any knowledge of the grammar of commands.
Long-horizon tasks have also been considered in one-shot imitation works~\cite{rocky,huang2018neural}. We consider a different setting where only demonstrations of primitives are available during meta-training, rather than demonstrations of full-length compound tasks.

A number of prior works have aimed to decompose demonstrations into primitives or skills, e.g. using change point detection~\cite{konidaris2012robot}, latent variable models~\cite{butterfield2010learning,niekum2013incremental,hausman2017multi}, mixture of experts~\cite{grollman2010incremental}, or transition state clustering~\cite{murali2016tsc}. Like these prior works, we build upon the notion that each primitive has its own policy, aiming to construct a sequence of policies, each with a learned termination condition. Similar to Manschitz et al.~\cite{manschitz2015probabilistic}, we learn the termination condition by predicting the phase of a primitive skill.
Unlike these approaches, we consider the problem of learning an end-to-end visuomotor policy (pixels to end-effector velocities) from a single video of a human performing a task, while leveraging visual demonstration data from primitives performed with previous objects.

\newcommand{\loss}{\mathcal{L}}
\newcommand{\data}{\mathcal{D}}
\newcommand{\task}{\mathcal{T}}
\newcommand{\humandata}{\data^h}
\newcommand{\humandemo}{\mathbf{d}^h}
\newcommand{\demo}{\mathbf{d}}
\newcommand{\robotdata}{\data^r}
\newcommand{\robotdemo}{\mathbf{d}^r}
\newcommand{\learnedloss}{\loss_\psi}
\newcommand{\bcloss}{\loss_\text{BC}}

\section{Preliminaries}
\label{sec:prelim}

To learn how to learn and compose primitive skills into multi-stage tasks, we need to first learn how to learn a primitive skill by imitating a human demonstration. To do so, our approach uses prior work on domain-adaptive meta-learning (DAML)~\cite{daml_rss} that learns how to infer a policy from a single human demonstration. DAML is an extension of the model-agnostic meta-learning algorithm (MAML)~\cite{maml}. In this section, we present the overview meta-learning problem, discuss both MAML and DAML, and introduce notation.


A primary goal of many meta-learning algorithms is to learn new tasks with a small amount of data. To achieve this, these methods learn to efficiently learn many meta-training tasks, such that, when faced with a new meta-test task, it can be learned efficiently. Note that meta-learning algorithms assume that the meta-training and meta-test tasks are sampled from the same distribution $p(\task)$, and that there exists a common structure among tasks, such that learning this structure can lead to fast learning of new tasks (referred as few-shot generalization). Hence, meta-learning corresponds to structure learning. 

MAML aims to learn the shared structure across tasks by learning parameters of a deep network such that one or a few steps of gradient descent leads to effective generalization to new tasks. We will use $\theta$ to denote the initial model parameters and  $\loss(\theta, \data)$ to denote the loss function of a supervised learner, where $\data_\task$ denotes labeled data for task $\task$. 
During meta-training, MAML samples a task $\task$ and datapoints from $\data_\task$, which are randomly partitioned into two sets,
$\data^\text{tr}_\task$ and $\data^\text{val}_\task$. 
We will assume that there are $K$ examples in $\data^\text{tr}_\task$.
MAML optimizes for model parameters $\theta$ such that one or a few gradient steps on $\data^\text{tr}_\task$ results in good performance on $\data^\text{val}_\task$. Concretely, the MAML objective is:
\vspace{-0.05cm}
\begin{align*}
    &\min_\theta  \sum_{\task \sim p(\task)} \loss(\theta-\alpha \nabla_\theta \loss(\theta, \data^\text{tr}_\task), \data^\text{val}_\task) \\
    &= \min_\theta \sum_{\task \sim p(\task)} \loss(\phi_\task, \data^\text{val}_\task).
    \vspace{-0.1cm}
\end{align*}
where $\phi_\task$ corresponds to the updated parameters and $\alpha$ is a step size of the gradient update. Moving forward, we will refer to the inner loss function as the \emph{adaptation objective} and the outer objective as the \emph{meta-objective}. At meta-test time, in order to infer the updated parameters for a new, held-out task $\task_\text{test}$, MAML runs gradient descent with respect to $\theta$ using $K$ examples drawn from $\task_\text{test}$:
\vspace{-0.1cm}
$$
\phi_{\task_\text{test}} = \theta - \alpha \nabla_\theta \loss(\theta, \data^{\text{tr}}_{\task_\text{test}}).
\vspace{-0.1cm}
$$

\begin{figure*}[t]
\setlength{\unitlength}{0.5\columnwidth}
\begin{picture}(1.99,1.15) \linethickness{0.5pt}
\put(0.1,-0.3){\includegraphics[width=1.95\columnwidth]{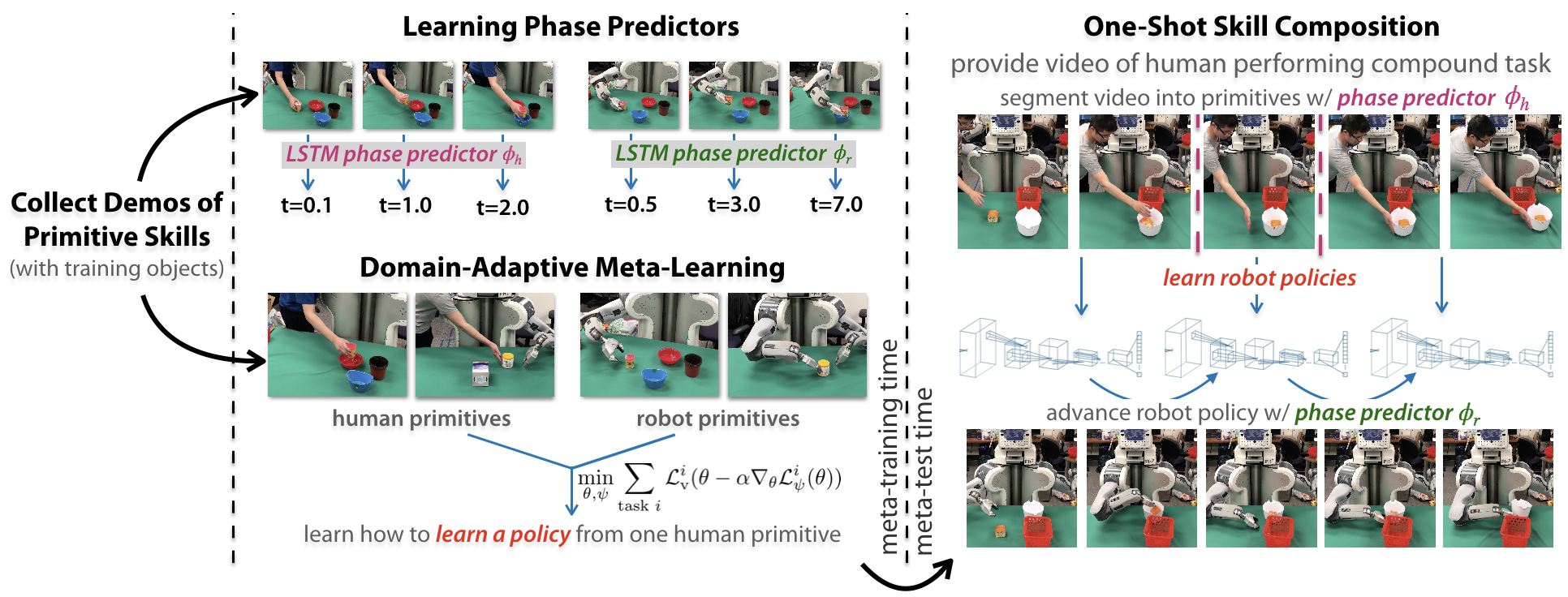}}
\end{picture}
\vspace{0.9cm}
\caption{
\label{fig:diagram}
After learning a phase predictor and meta-learning with human and robot demonstration primitives, the robot temporally segments a human demonstration (just a video) of individual primitives and learns to perform each segmented primitive sequentially.}
\vspace{-0.14in}
\end{figure*}

DAML applied the MAML algorithm to the domain-adaptive one-shot imitation learning setting; DAML aims to learn how to learn from a video of a human, using teleoperated demonstrations for evaluating the meta-objective. Essentially, DAML learns to translate from a video of a human performing a task to a policy that performs that task.
Unlike the standard supervised meta-learning setting, a video of a human is a form of weak supervision -- it contains enough information to communicate the task, but does not contain direct label supervision.
To allow a robot to learn from a video of a human and handle the domain shifts between the human and the robot, DAML additionally learns the \emph{adaptation objective} denoted as $\learnedloss$ along with the initial parameters $\theta$. The meta-objective is a mean-squared error behavioral cloning loss denoted as $\bcloss$. $\learnedloss$ can be viewed as a learned critic such that running gradient descent on $\learnedloss$ can produce effective adaptation. Specifically, the DAML meta-objective can be formulated as follows:
$$
\min_{\theta, \psi} \sum_{\task \sim p(\task)} \sum_{\humandemo \in \humandata_\task} \sum_{\robotdemo \in \robotdata_\task} \bcloss( \theta - \alpha \nabla_\theta \learnedloss(\theta, \humandemo), \robotdemo). 
$$
where $\humandemo$ and $\robotdemo$ are a human and robot demonstration respectively. In prior work with DAML, different tasks corresponded to different objects, hence having the robot learn to perform previous skills with novel objects. In this work, different meta-learning tasks will correspond to different primitives performed with different sets of objects.
Next, we present our approach for one-shot visual skill composition.


\newcommand{\obs}{\mathbf{o}}
\newcommand{\state}{\mathbf{s}}
\newcommand{\action}{\mathbf{a}}
\newcommand{\primitive}{\mathcal{P}}
\newcommand{\humanphasepred}{\phi_h}
\newcommand{\robotphasepred}{\phi_r}

\section{One-Shot Visual Skill Composition}
\label{sec:method}

Our goal is for a robot to learn to perform a variety of multi-stage tasks by ``watching'' a video of a human perform each task. To do so, the robot needs to have some form of prior knowledge or experience, which we will obtain
by using demonstration data of primitive subtasks performed with other objects. After meta-training, at meta-test time, the robot is provided with a single video of a human performing a multi-stage task, and is tasked with learning to perform the same multi-stage task in new settings where the objects have moved into different starting configurations.
In this section, we define the problem assumptions and present our approach for tackling this problem setting.

\subsection{Problem Setting and Overview}

Concretely, during an initial meta-training phase,
we provide a dataset of primitive demonstrations. For each subtask with a particular set of objects (which we refer to as a primitive $\primitive_k$), we provide multiple human demonstrations, $\{\humandemo_i\}_k$ and multiple robot demonstrations $\{\robotdemo_j\}_k$. We define a demonstration performed by a human $\humandemo_k$ to be a sequence of images $\obs^h_1, \dots, \obs^h_{T_i}$ of a human performing $\primitive_k$, and a robot demonstration $\robotdemo_k$ to be a sequence of images and actions $\obs^r_1, \action^r_1 \dots, \obs^r_{T_j}, \action^r_{T_j}$ of a robot performing the same primitive. Note that demonstrations have potentially different time horizons, namely it may be that $T_i \neq T_j$.
The human and robot demonstration need to have a correspondence
based on the objects used and primitive performed, but do not need to be aligned in any way, nor be executed with the same speed, nor have the same object positions.

After meta-training, the robot is provided with a human demonstration $\humandemo$ of a compound task consisting of multiple primitives in sequence. The sequence of primitives seen in this video at meta-test time may involve novel objects and novel configurations thereof, though we assume that the general families of subtasks in the human demonstration (e.g., pushing, grasping, etc) are seen in the meta-training data. However, the meta-training data does not contain any composition of primitives into multi-stage tasks. After observing the provided human video, the robot is tasked with identifying primitives, learning policies for each primitive in the human's demonstration, and composing the policies together to perform the task.

In our approach, we use the meta-training
data to have the robot learn how to learn an individual subtask from a video demonstration, in essence, translating a video demonstration into a policy for the subtask. Additionally, the robot needs to learn how to identify and compose learned primitives into multi-stage skills. To do so, we propose to train two models that can identify the temporal progress, or \emph{phase}, of any human or robot primitive respectively, which can be used for segmentation of human demonstrations and for terminating a primitive being performed by the robot to move onto the next subtask. An overview of our method is in Figure~\ref{fig:diagram}. In the remainder of this section, we discuss how we can learn and compose primitives for completing compound tasks from single demonstrations, and how our two phase predictors can be trained.




\begin{algorithm}[t]
    \caption{One-Shot Composition of Primitives}
    \label{alg:composition}
\begin{algorithmic}
\REQUIRE meta-learned initial parameters $\theta$, and adaptation objective $\loss_\psi$
\REQUIRE human and robot phase predictors $\phi_h$ and $\phi_r$
\REQUIRE A human video for a compound task $\obs^h_1, \dots, \obs^h_T$
\STATE \textit{\# Decompose human demonstration into primitives}
\STATE Initialize predicted primitives $\primitive$ = $\{\}$ and $t = 1, t' = 1$
\WHILE{$t < T$}
    \IF{$\phi_h(\obs^h_{t':t}) > 1 - \epsilon$}
    \STATE Append $\demo^h = \obs^h_{t':t}$ to $\primitive$
    \STATE $t' = t + 1$
    \ENDIF
    \STATE $t = t + 1$
\ENDWHILE
\STATE \textit{\# Compute and compose policies for each primitive}
\STATE Initialize $t=1$, $t'=1$ observe $\obs^r_1$
\FOR{$\demo^h_i$ in $\primitive$}
    \STATE Compute policy parameters $\phi_i = \theta - \alpha \nabla_\theta \learnedloss (\theta, \demo^h_i)$
    \WHILE{$\robotphasepred(\obs^r_{t':t}) < 1 - \epsilon$}
    \STATE Take one step, executing $\pi_{\phi_i}(\obs^r_t)$ and getting $\obs^r_{t+1}$
    \STATE $t = t + 1$
    \ENDWHILE
    \STATE $t' = t$
\ENDFOR
\end{algorithmic}
\end{algorithm}

\subsection{One-Shot Composition of Primitives}
\label{sec:one_shot_composition}


During meta-training, we train a human phase predictor $\phi_h$ and a robot phase predictor $\phi_r$ (as described in the following section), as well as a DAML one-shot learner $\pi_\theta$, which can learn primitive-specific robot policies $\pi_{\phi_i}$ from videos of humans performing primitives, $\humandemo_i$.
At meta-test time, the robot is provided with a video of a human completing a multi-stage task, $\obs^h_1, \dots, \obs^h_{T}$.

We first need to decompose the multi-stage human demonstration into individual primitives. We can do so by using the human phase predictor. In particular, we feed the demonstration, frame by frame, into $\humanphasepred$ until $\humanphasepred(\obs^h_{1:t}) > 1 - \epsilon$, indicating the end of the current primitive, $\humandemo=\obs^h_1, ..., \obs^h_t$. Then, we repeat this process starting from the following timestep $t+1$, to iteratively determine the endpoint of each segment. At the end of this process, we are left with a sequence of primitive demonstrations, $\humandemo_1, \humandemo_2, ...$, which we translate into policies $\pi_{\phi_1}, \pi_{\phi_2}, ...$ using the one-shot imitator $\pi_\theta$. To compose these policies together, we roll-out each policy sequentially, passing the observed images $\obs_t$ into the robot phase predictor and advancing the policy when $\robotphasepred(\obs^r_{1:t}) > 1 - \epsilon$. We detail this meta-test time process in Algorithm~\ref{alg:composition}.



\subsection{Primitive Phase Prediction}
\label{sec:phase_pred}

In order to segment a video of a compound skill into a set of primitives, many prior works train on a dataset consisting of multi-stage task demonstrations and corresponding labels that indicate the primitive at each time step. We assume a dataset of demonstrations of primitives, but without any label for the particular primitive that was executed in that segment, and where the primitive demonstrations contain different objects than those seen at meta-test time.
As a result, we don't need a dataset of demonstrations of temporally extended skills during training, which can reduce the burden on humans to collect many long demonstrations. However, we still need to learn both (a) how to segment human demonstrations of compound tasks, in order to learn a sequence of policies that perform the task shown at meta-test time, and (b) when to transition from one learned policy to the next. We propose a single approach that can be used to tackle both of these problems: predicting the \emph{phase} of a primitive. In particular, given a video of a partial demonstration or execution of any primitive, we aim to predict how much temporal progress has been made towards the completion of that primitive. This form of model can be used both to segment a video of a compound task and to identify when to switch to a subsequent policy.


We train two phase predictor models, one on human demonstration data and one on robot demonstration data. Other than the data, both models have identical form and are trained in the same way. As discussed previously, the meta-training dataset
is composed of human and robot demonstrations of a set of primitives. To construct supervision for training the phase predictors, we can compute the phase of particular partial video demonstration $\obs_{1:t}$ as $\frac{t}{T}$ where $T=|\demo|$ is the length of the full demonstration $\demo$, which varies across demonstrations. The phase information provides label supervision that indicates how much of the primitive has been completed. Each phase predictor is trained to take as input a partial demonstration, i.e. the first $t$ frames of a demonstration, and output the phase.


Formally, we denote $\humanphasepred$ as the human phase predictor, which takes as input $\obs^h_{i, 1:t}=\obs^h_{i,1}, ..., \obs^h_{i,t}$ and regresses $\humanphasepred(\obs^h_{i, 1:t})$ to $\frac{t}{T_i}$. Similarly, $\robotphasepred$ is defined to be the robot phase predictor, regressing $\robotphasepred(\obs^r_{j, 1:t})$ to $\frac{t}{T_j}$, where the partial demonstration $\obs^r_{j, 1:t} = \obs^r_{j,1}, ..., \obs^r_{j,t}$ only uses image observations for simplicity. To handle variable length sequences as input, both models $\humanphasepred$ and $\robotphasepred$, are represented using recurrent neural networks. Both networks are trained using a mean-squared error objective
$
\sum_i \sum_{t=1}^{T_i} \left|\left| \phi(\demo_{i,1:t}) - \frac{t}{T_i} \right|\right|^2
$
and the Adam optimizer~\cite{adam}.
The phase predictors use $5$ convolution layers with $64$ $3 \times 3$ filters, followed by an LSTM with $50$ hidden units and a linear layer to output the phase. The first convolution layer is initialized using pretrained weights from VGG-16. We use the swish non-linearity~\cite{swish}, layer normalization~\cite{layernorm}, and dropout~\cite{dropout} with probability $0.5$ at each convolution layer and the LSTM.


\section{Experiments}
\label{sec:result}

The goal of our experiments is to determine whether our approach can enable a robot to learn to perform compound vision-based tasks from a single video demonstration of a human performing that task, composing primitives on the fly and generalizing to new objects. Note that this is an exceptionally challenging task: in contrast to prior work on compound task learning~\cite{manschitz2015learning,maryland,xu2017neural,shiarlis2018taco}, the robot must perform the task entirely end-to-end from RGB images and receives only a single unsegmented video demonstration showing a person performing the task. The meta-training data does not contain instances of the same objects that are seen in the test-time behaviors, requiring the robot to adapt its policy to each of the objects from the prior learned during meta-training. Each task requires sequencing multiple primitives, such that no single policy can perform the entire task alone. Success on this problem requires simultaneously interpreting the human demonstration to determine which objects matter for the task and how they should be manipulated, adapting the policy multiple times to the multiple stages of the task, and executing these adapted policies in the right sequence. 

Our experimental set-ups involve pick-and-place primitives, push primitives, and reach primitives. Hence, our method is capable of composing these primitives in a variety of ways to form many different compound tasks. In our evaluation, we focus on two different sets of meta-test tasks:

\paragraph{Simulated order fulfillment} A simulated Sawyer robot must learn to pick and place a particular set of novel objects into a bin and push the bin to a specified location. 

\paragraph{PR2 kitchen serving} In this setting, the PR2 must grasp an object, place it into the correct bowl or platter, and push one of the platters or bowls to the robot's left.

To our knowledge, no prior work proposes an approach that aims to solve the problem that we consider. Because we do not assume access to demonstrations of compound tasks during meta-training, it is not suitable to directly compare to direct one-shot imitation on compound tasks. Even if there are many compound task demonstrations for meta-training, no prior work has demonstrated one-shot imitation learning of temporally extended tasks from raw pixels.
Consequently, we compare only to ablations of our method to better understand the importance of using phase prediction and using DAML over alternative options. For understanding the former, we compare to a simple alternative to using phase prediction: learning policies for every fixed-length segment of the human demonstration and advancing the policy at every timestep. This `sliding window' approach still leverages one-shot imitation, but makes two simplifying assumptions: (a) that fixed-length windows are an appropriate representation of primitives, and (b) that the human demonstration time and robot execution time are equal. This latter assumption will be true in our simulated experiments, where the provided demonstrations are from a robot, but may be easily violated in the real world. To study the importance of using DAML, we compare to using an LSTM-based meta-learner, akin to model of Duan et al.~\cite{rocky} but using visual inputs.
All methods use Cartesian end-effector control. 
The value of $\epsilon$ is set to $0.03$, which was found to work well on validation tasks. For video results, see the project website\footnote{\url{https://sites.google.com/view/one-shot-hil}}.




\begin{figure*}[t]
\setlength{\unitlength}{0.5\columnwidth}
\begin{picture}(1.99,1.5) \linethickness{0.5pt}
\put(0.0,0.7){\includegraphics[width=1.0\columnwidth]{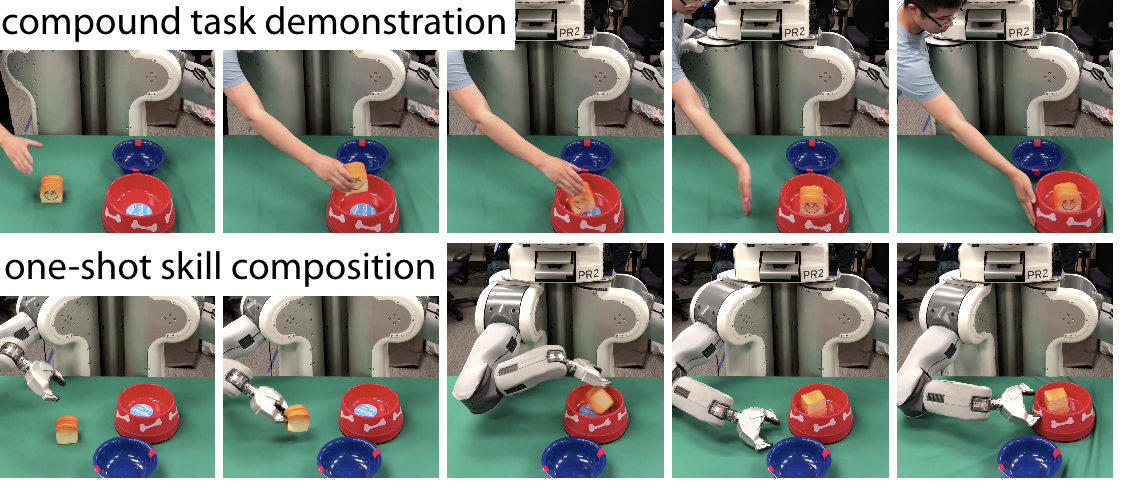}}
\put(2.1,0.7){\includegraphics[width=1.0\columnwidth]{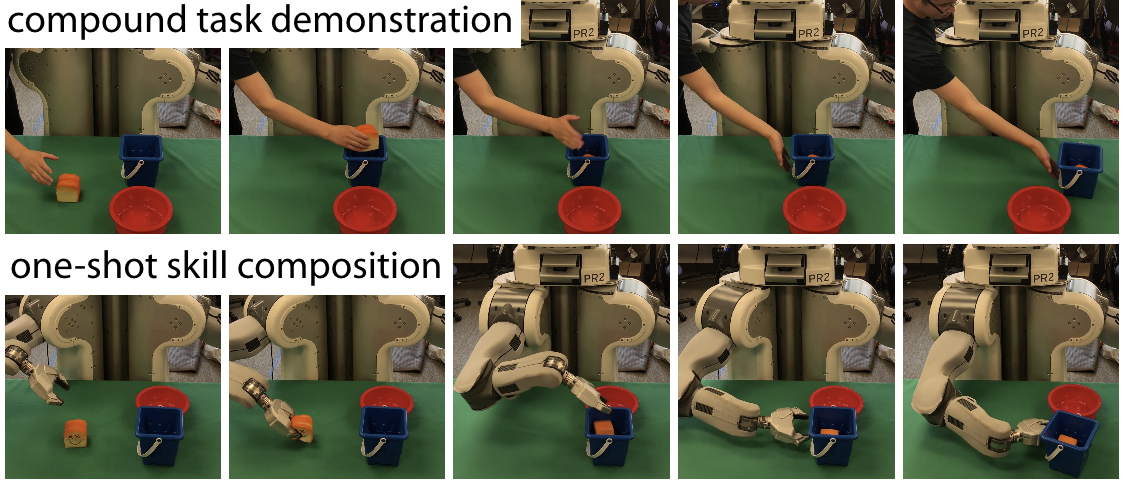}}
\put(0.0,0.0){\includegraphics[width=1.0\columnwidth]{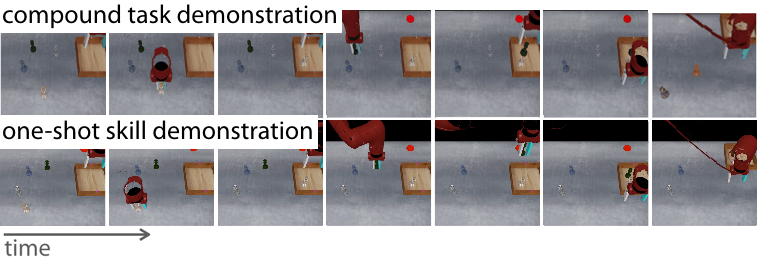}}
\put(2.1,0.0){\includegraphics[width=1.0\columnwidth]{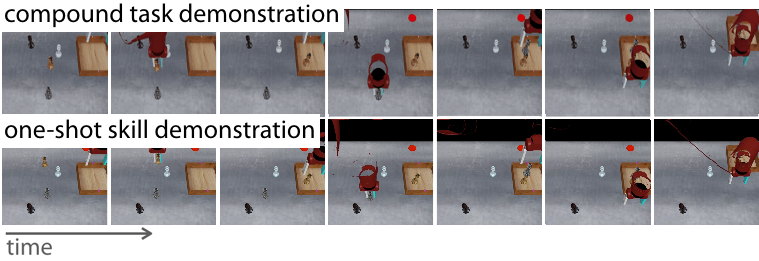}}
\end{picture}
\vspace{-0.3cm}
\caption{
\label{fig:qual}
Qualitative results of the experiments on the physical PR2 (top) and simulated Sawyer (bottom). The top right shows an example failure case, where the robot successfully picks and places the object into the correct container, but incorrectly pushes between the two objects. The other examples all illustrate successful learning and skill composition of the demonstrated compound task from a single demonstration.}
\end{figure*}



\subsection{Simulated Order Fulfillment}

We first evaluate our method on a range of simulated order fulfillment tasks using a Sawyer robot arm in the MuJoCo physics engine~\cite{mujoco}, as illustrated in Figure~\ref{fig:qual}. 
Across tasks, the particular objects and number of objects to be put in the bin vary. Success is defined as having the correct objects in the bin and the bin at the target. 
In this experiment, we consider a simplified setting, where a typical demonstration (with both images and actions) is available at meta-test time. The next section with physical robot experiments will consider learning from human videos.

There are two types of primitive demonstrations in the training dataset: picking and placing a single object into the bin, and pushing the bin to the goal. In lieu of a simulated human demonstrator, we optimize an expert policy for pick and place using proximal policy optimization (PPO)~\cite{schulman2017proximal} and script an expert pushing policy. The PPO expert policy uses priveleged low-level state information (i.e. the position of the target object) rather than vision, enabling us to train a single expert policy across all pick \& place primitives.
To gather pick \& place primitives for the training dataset, we sample one target object and two or three distractors from a set of $37$ types of textures where each type contains roughly $150$ different textures (for a total of around $5500$ textures). We also randomize the positions of the target and distractors to form $8$ different demonstrations for each primitive. We use $1800$ different pick-and-place primitive subtasks, along with $56$ bin pushing demonstrations, for meta-learning and training the phase predictors.

We represent the DAML network with $4$ convolution layers with $24$ $5 \times 5$ filters, followed by $3$ fully-connected layers with $200$ hidden units.
We use linear adaptive objectives~\cite{daml_rss} for both actions and the gripper pose, and a step size $\alpha=0.05$. The meta-objective is the same as prior work, as is the LSTM meta-learner architecture~\cite{daml_rss}.

We evaluate each approach using one-object and two-object order fulfillment tasks. For each multi-stage meta-test task, we generate one visual demonstration by temporally concatenating expert demonstrations of the two or three primitives involved in the task, as illustrated in Figure~\ref{fig:qual}.
We compute success averaged over 10 tasks of each type and 3 trials per task. The results, shown in Table~\ref{tbl:sim_results}, indicate that both gradient-based meta-learning and phase prediction are essential for good performance. We also observe that performance drops as the task becomes longer, indicative of compounding errors that are known to be a problem when using behavioral cloning based approaches~\cite{dagger}. Generally, we find that the nearly all failure cases are caused by the one-shot imitation learner, primarily related to grasping, which implicitly requires precise visual object localization and precise control. Hence, future advances in one-shot visual imitation learning will likely lead to improvements in our approach as well.

\begin{table}[h]
    \begin{center}
    \vspace{-0.3cm}
    \begin{tabular}{|l|c|c|}
    \hline
      & 1 object &  2 objects  \\
      \hline
      sliding window (no phase prediction) &   50.0\% & 16.7\%     \\ 
      \hline
      LSTM one-shot learner (no DAML) & 0.0\% & 0.0\% \\
      \hline
      \textbf{one-shot skill composition (ours)} &  \textbf{73.3\%} & \textbf{46.7\%} \\
      \hline
    \end{tabular}
    \end{center}
    \vspace{-0.2cm}
    \caption{One-shot success rate of simulated Sawyer robot performing order fulfillment from a single demonstration with comparisons
    }
    \vspace{-0.8cm}
    \label{tbl:sim_results}
\end{table}

\subsection{PR2 Kitchen Serving}

\begin{figure}[t]
\setlength{\unitlength}{0.5\columnwidth}
\begin{picture}(0.99,0.75) \linethickness{0.5pt}
\put(0.425,0.175){\includegraphics[width=0.25\columnwidth]{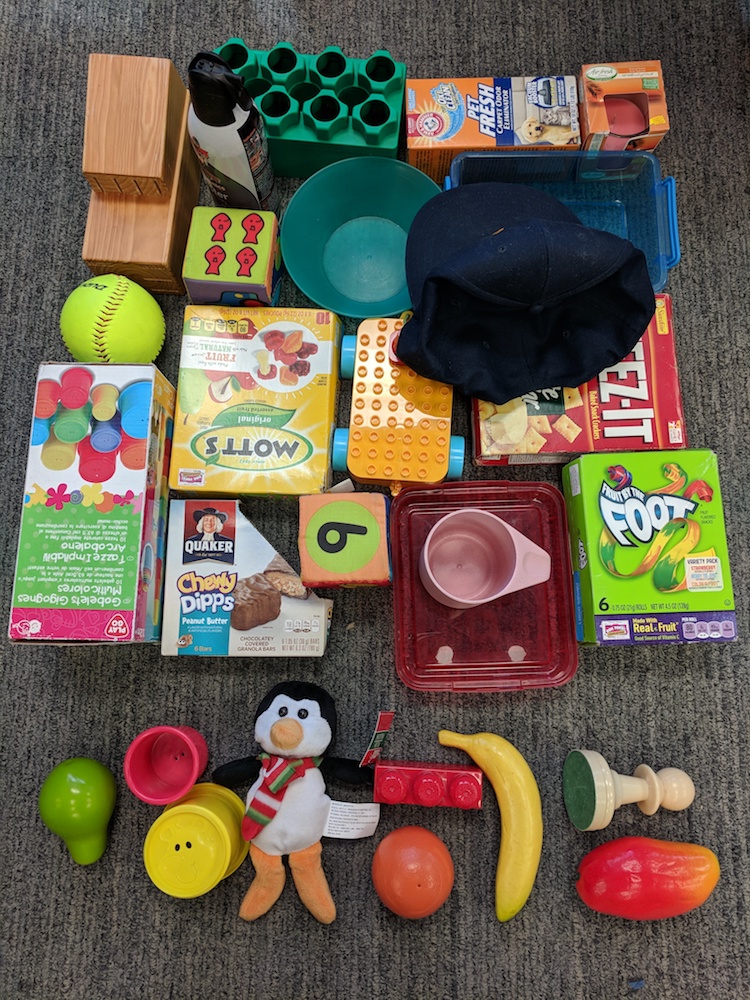}}
\put(0.975, 0.175){\includegraphics[width=0.25\columnwidth]{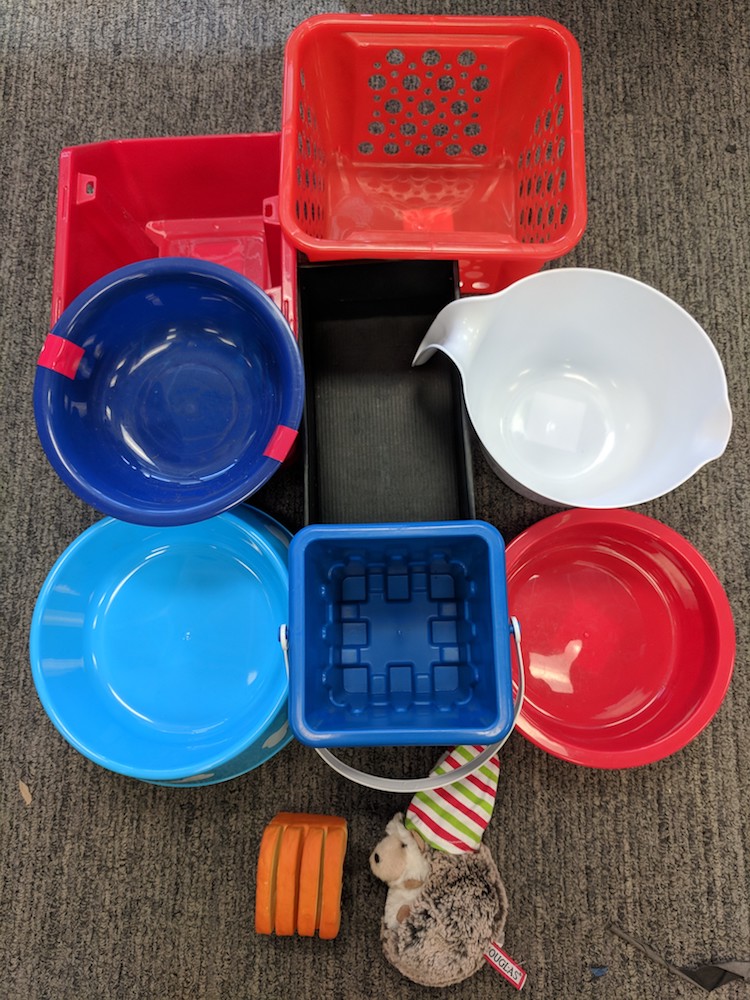}}
\end{picture}
\vspace{-1.0cm}
\caption{
\label{fig:objects}
A subset of the meta-training objects (left) and meta-test objects (right) used in the PR2 experiments. The top of each picture shows placing and pushing target objects and the bottom shows picked objects.
}
\vspace{-1.0cm}
\end{figure}

In our second experiment on a physical PR2 robot, we collect primitive demonstrations by (a) using the dataset from \cite{daml_rss}, which contains $1293$, $640$, $1008$, and $600$ robot demonstrations for placing, pushing, pick-and-place, and (b) primitives that transition from placing to pushing, plus an equal number of human demonstrations for the above four primitives. 
We focus on two particular tasks for this experiment: picking an object, placing it into a particular container, and then pushing either the container with the object or a different container toward a goal position. 
The success metric for this task is that the object is placed into the container and the container is pushed toward the robot's left gripper. 

We follow the same DAML architectures as in prior work~\cite{daml_rss} with two small differences. Instead of fully-connected layers, the model uses temporal convolution layers; and rather than using a mixture density network at the output, the architecture uses a discretized action space and a cross-entropy loss with each action dimension independently discretized over $50$ bins.

For evaluation, we use 10 novel bowls and containers and 5 novel items to be `served' (see Figure~\ref{fig:objects}). For each task, we collect a single human demonstration and evaluate two trials of the robot's task execution, as illustrated in Figure~\ref{fig:qual}. 
We summarize the results in Table \ref{tbl:real_results}. 
Since this is a challenging task as the robot needs to complete a sequence of control primitives and any small drift during imitation could lead to failure, our approach can only succeed $10$ out of $20$ trials when pick-and-placing the object into the target the container and pushing the correspond container, and $7$ out of $20$ trials when pushing the other container without the placed object. However, the na\"ive sliding window approach never successfully completes the entire sequence of primitives, indicating that phase prediction is an important aspect of our approach. As before, we observe that most of the failure cases of our approach hinge on the one-shot imitation learning failing to perform the primitive in the segmented video (see example in Figure~\ref{fig:qual}), suggesting that future advances in one-shot imitation from human videos will lead to improved performance.

\begin{table}[h]
    \begin{center}
    \vspace{-0.5cm}
    \begin{tabular}{|l|c|c|}
    \hline
      & same target &  different target\\
     \hline
      sliding window (no phase prediction) &   0/20 & 0/20 \\ 
      \hline
      \textbf{one-shot skill composition (ours)} &   \textbf{10/20} & \textbf{7/20} \\
      \hline
    \end{tabular}
    \end{center}
    \vspace{-0.2cm}
    \caption{Success rate of PR2 robot performing pick \& place then push task from a video of a human. Evaluated using novel, unseen objects.
    }
    \vspace{-1.0cm}
    \label{tbl:real_results}
\end{table}


\section{Discussion}
\label{sec:conclusion}

We presented an approach for one-shot learning and composition policies for achieving compound, multi-stage tasks from raw pixel inputs based on a single video of a human performing the task. Our approach leverages demonstrations from previous primitive skills in order to learn to identify the end of a primitive and to meta-learn policies for primitives. At meta-test time, our approach learns multi-stage tasks by decomposing a human demonstration into primitives, learning policies for each primitive, and composing the policies online to execute the full compound task. Our experiments indicated that our approach can successfully learn and compose convolutional neural network policies for order fulfillment tasks on a simulated Sawyer arm and a real PR2 arm, all from raw pixel input. 

In future work, we hope to improve upon the performance of our approach. To do so, it will be important to improve the performance of one-shot imitation learning methods (a  subcomponent of our approach) and potentially incorporate reinforcement learning or other forms of online feedback, such as DAgger~\cite{dagger}, to overcome compounding errors. Another interesting direction for future work is to consider more unstructured demonstration data, such as human and robot demonstrations of temporally-extended tasks that are from distinct sources. Unlike many prior works that have considered the related problem of unsupervised segmentation of compound task demonstrations into primitives, this problem is significantly more challenging, since it also involves learning a rough alignment between skills performed by humans and those performed by robots. But if possible, the method would provide the ability to scale to large, unlabeled datasets and may remove the need for an expert to decide the set of skills that constitute primitives during training.



\section*{Acknowledgements} We thank Kate Rakelly and Michael Chang for helpful comments on an early version of this manuscript. This work was supported by Berkeley DeepDrive, the National Science Foundation through IIS-1700696 and a Graduate Research Fellowship. We also acknowledge NVIDIA, Amazon, and Google for equipment and computing support.

\bibliographystyle{IEEEtran}
\bibliography{references}

\end{document}